\documentclass[conference]{IEEEtran}
\ifCLASSINFOpdf
   \usepackage[pdftex]{graphicx}
\else
   \usepackage[dvips]{graphicx}
\fi
\usepackage{amsmath}
\ifCLASSOPTIONcompsoc
  \usepackage[caption=false,font=normalsize,labelfont=sf,textfont=sf]{subfig}
\else
  \usepackage[caption=false,font=footnotesize]{subfig}
\fi
\usepackage{url}

\usepackage{verbatim}

\usepackage{algorithm}
\usepackage{algpseudocode}

\algnewcommand{\IIf}[1]{\State\algorithmicif\ #1\ \algorithmicthen}
\algnewcommand{\IIfa}[1]{\hspace{\algorithmicindent}\State\algorithmicif\ #1\ \algorithmicthen}
\algnewcommand{\EndIIf}{\unskip\ \algorithmicend\ \algorithmicif}

\let\AState\State
\let\State\relax

\usepackage{color}

\usepackage[utf8]{inputenc}
\usepackage[T1]{fontenc}

\begin{document}
%
\title{Text-based Adventures of the Golovin AI Agent}

\author{
\IEEEauthorblockN{Bartosz Kostka, Jaros{\l}aw Kwiecie{\'n}, Jakub Kowalski, Pawe{\l} Rychlikowski}
\IEEEauthorblockA{Institute of Computer Science\\
University of Wroc{\l}aw, Poland\\
Email: \{bartosz.kostka,jaroslaw.kwiecien\}@stud.cs.uni.wroc.pl, \{jko,prych\}@cs.uni.wroc.pl}


}


%


\maketitle

\begin{abstract}
The domain of text-based adventure games has been recently established as a new challenge of creating the agent that is both able to understand natural language, and acts intelligently in text-described environments.

In this paper, we present our approach to tackle the problem. Our agent, named Golovin, takes advantage of the limited game domain. We use genre-related corpora (including fantasy books and decompiled games) to create language models suitable to this domain. Moreover, we embed mechanisms that allow us to specify, and separately handle, important tasks as fighting opponents, managing inventory, and navigating on the game map.

We validated usefulness of these mechanisms, measuring agent's performance on the set of 50 interactive fiction games. Finally, we show that our agent plays on a level comparable to the winner of the last year Text-Based Adventure AI Competition.

%
%

\end{abstract}




%
\IEEEpeerreviewmaketitle





\section{Introduction}

The standard approach to develop an agent playing a given game is to analyze the game rules, choose an appropriate AI technique, and incrementally increase the agent's performance by exploiting these rules, utilizing domain-dependent features and fixing unwanted behaviors. This strategy allowed to beat the single games which were set as the milestones for the AI development: Chess \cite{Campbell2002Deep} and Go \cite{Silver2016Mastering}.

An alternative approach called  General Game Playing (GGP), operating on a higher level of abstraction has recently gained in popularity. Its goal is to develop an agent that can play any previously unseen game without human intervention. Equivalently, we can say that the game is one of the agent's inputs \cite{Genesereth2014General}.

Currently, there are two main, well-established GGP domains providing their own game specification languages and competitions \cite{Genesereth2005General}. The first one is the Stanford's GGP, emerged in 2005 and it is based on the Game Description Language (GDL), which can describe all finite, turn-based, deterministic games with full information \cite{Love2008General}, and its extensions (GDL-II \cite{Thielscher2010AGeneral} and rtGDL \cite{Kowalski2016Towards}).

The second one is the General Video Game AI framework (GVGAI) from 2014, which focuses on arcade video games \cite{Perez2015The2014}. In contrast to Stanford's GGP agents are provided with the forward game model instead of the game rules. The domain is more restrictive but the associated competition provides multiple tracks, including procedural content generation challenges \cite{Perez2016General,Khalifa2016General}.

What the above-mentioned approaches have in common is usually a well-defined game state the agent is dealing with. It contains the available information about the state (which may be partially-observable), legal moves, and some kind of scoring function (at least for the endgame states). Even in a GGP case, the set of available moves is known to the agent, and the state is provided using some higher-level structure (logic predicates or state observations).

In contrast, the recently proposed  Text-Based Adventure AI Competition, held during the IEEE Conference on Computational Intelligence and Games (CIG) in 2016, provides a new kind of challenge by putting more emphasis on interaction with the game environment. The agent has access only to the natural language description about his surroundings and effects of his actions. 

Thus, to play successfully, it has to analyze the given descriptions and extract high-level features of the game state by itself. Moreover, the set of possible actions, which are also expected to be in the natural language, is not available and an agent has to deduct it from his knowledge about the game's world and the current state.

In some sense, this approach is coherent with the experiments on learning Atari 2600 games using the Arcade Learning Environment (ALE), where the agent's inputs were only raw screen capture and a score counter \cite{Mnih2015HumanLevel,Kelly2017Emergent}. Although in that scenario the set of possible commands is known in advance.

The bar for text-based adventure games challenge is set high -- agents should be able to play any interactive fiction (IF) game, developed by humans for the humans. Such environment, at least in theory, requires to actually understand the text in order to act, so completing this task in its full spectrum, means building a strong AI.

Although some approaches tackling similar problems exist since early 2000s (\cite{depristo2001being,amir2002adventure}), we are still at the entry point for this kind of problems, which are closely related to the general problem solving. However, recent successes of the machine learning techniques combined with the power of modern computers, give hope that some vital progress in the domain can be achieved.

We pick up the gauntlet, and in this work we present our autonomous agent that can successfully play many interactive fiction games. We took advantage of the specific game domain, and trained agent using matching sources: fantasy books and texts from decompiled IF games. Moreover, we embed some rpg-game-based mechanisms, that allow us to improve fighting opponents, managing hero's inventory, and navigating in the maze of games' locations.

We evaluated our agent on a set of 50 games, testing the influence of each specific component on the final score. Also, we tested our agent against the winner of the last year competition \cite{Fulda2017WhatCan}. The achieved results are comparable. Our agent scored better in 12 games and worse in 11 games.

The paper is organized as follows.  Section~II provides background for the domain of interactive fiction, Natural Language Processing (NLP), Text-Based Adventure AI Competition, and the related work. In Section~III, we presented detailed description of our agent. Section~IV contains the results of the performed experiments. Finally, in Section~V, we conclude and give perspective of the future research.

\section{Background}

\subsection{Interactive Fiction}

Interactive Fiction (IF), emerged in 1970s, is a domain of text-based adventure or role playing games, where the player uses text commands to control characters and influence the environment. One of the most famous example is the Zork series developed by the Infocom company. From the formal point of view, they are single player, non-deterministic games with imperfect information. IF genre is closely related to MUDs (Multi-User Dungeons), but (being single-player) more focused on plot and puzzles than fighting and interacting with other players. IF was popular up to late 1980s, where the graphical interfaces become available and, as much user friendlier, more popular. Nevertheless, new IF games are still created, and there are annual competitions for game authors (such as The Interactive Fiction Competition).

IF games usually (but not always) take place in some fantasy worlds. The space the character is traversing has a form of labyrinth consisting of so called rooms (which despite the name can be open areas like \emph{forest}). Entering the room, the game displays its description, and the player can interact with the objects and game characters it contains, or try to leave the room moving to some direction. However, reversing a movement direction (e.g.\ \emph{go south} $\leftrightarrow$ \emph{go north}) not necessarily returns the character to the previous room.

As a standard, the player character can collect objects from the world, store them in his equipment, and combine with other objects to achieve some effects on the environment (e.g.\ \emph{put the lamp and sword in the case}). Thus, many games require solving some kind of logical puzzle to push the action forward. After performing an action, the game describes its effect. Many available actions are viable, i.e.\ game engine understands them, but they are not required to solve the game, or even serve only for the player amusement.

Some of the games provide score to evaluate the player's progress, however the change in the score is often the result of a complex series of moves rather than quick ``frame to frame'' decisions, or the score is given only after the game's end. Other games do not contain any scoring function and the only output is win or lose.

\subsection{Playing Text-Based Games}

Although the challenge of playing text-based games was not take on often, there are several attempts described in the literature, mostly based on the MUD games rather than the classic IF games. 

Adventure games has been carefully revised as the field of study for the cognitive robotics in \cite{amir2002adventure}. First, the authors identify the features of ``tradition adventure game environment'' to point-out the specifics of the domain. Second, they enumerate and discuss existing challenges, including e.g.\ the need for commonsense knowledge (its learning, revising, organization, and using), gradually revealing state space and action space, vague goal specification and reward specification.

In \cite{depristo2001being}, the agent able to live and survive in an existing MUD game have been described. The authors used layered architecture: high level planning system consisting of reasoning engine based on hand-crafted logic trees, and a low level system responsible for sensing the environment, executing commands to fulfill the global plan, detecting and reacting in emergency situations. 

While not directly-related to playing algorithms, it is worth to note the usage of
computational linguistics and theorem proving to build an engine for playing text-based adventure games \cite{koller2004put}. Some of the challenges are similar for both tasks, as generating engine requires e.g.\ object identification (given user input and a state description) and understanding dependencies between the objects.

The approach focused on tracking the state of the world in text-based games, and translating it into the first-order logic, has been presented in \cite{hlubocky2004knowledge}. Proposed solution was able to efficiently  update agent's belief state from a sequence of actions and observations.


The extension of the above approach, presents the agent that can solve puzzle-like tasks in partially observable domain that is not known in advance, assuming actions are deterministic and without conditional effects \cite{Chang2006Goal}. It generates solutions by interleaving planning (based on the traditional logic-based approach) and execution phases. The correctness of the algorithm is formally proved.


Recently, an advanced MUD playing agent has been described in \cite{Narasimhan2015Language}. Its architecture consists of two modules. First, responsible for converting textual descriptions to state representation is based on the Long Short-term Memory (LSTM) networks \cite{Hochreiter1997Long}. Second, uses Deep Q-Networks \cite{Mnih2015HumanLevel} to learn approximated evaluations for each action in a given state. Provided results show that the agent is able to to successfully complete quests in small, and even medium size, games.

\subsection{Natural Language Processing}

Natural Language Processing is present in the history of computers almost from the very beginning. Alan Turing in his famous paper \cite{turing1950computing} state (approximately) that ``exhibit intelligent behavior'' means ``understand natural language and use it properly in conversations with human being''. So, since 1950 Turing test is the way of checking whether computer has reached strong AI capability. 

First natural language processing systems were rule based. Thanks to the growing amount of text data and increase of the computer power, during last decades one can observe the shift towards the data driven approaches (statistical or machine learning). Nowadays, NLP very often is done ``almost from scratch'', as it was done if \cite{DBLP:journals/corr/abs-1103-0398} where the authors have used neural network in order to solve many NLP tasks, including part-of-speech tagging, named entity recognition and semantic role labeling. The base for this was the neural language model. Moreover, this systems produced (as a side effect) for every word in a vocabulary a dense vector which reflected word properties. This vectors are called word embeddings and can be obtained in many ways. One of the most popular is the one proposed in  \cite{DBLP:journals/corr/abs-1301-3781} that uses very simple, linear language model and is suitable to large collections of texts. 

Language models allow to compute probability of the sentence treated as a sequence of items (characters, morphemes or words). This task was traditionally done using Markov models (with some smoothing procedures, see \cite{Chen:1996:ESS:981863.981904}). Since predicting current words is often dependent on the long part of history, Markov models (which, by definition, looks only small numbers of words behind) are outperformed by the modern methods that can model long distance dependencies. This methods use recursive (deep) neural networks, often augmented with some memory. 

We will use both word embeddings (to model words similarity) and LSTM neural networks \cite{Bengio03aneural} with attention mechanism (see \cite{DBLP:journals/corr/RaffelE15} and \cite{DBLP:journals/corr/ChengDL16}).
 We are particularly interested in the information given from the attention mechanism,
which allows us to estimate how important is each word, when we try to predict the next word in the text.

\subsection{The Text-Based Adventure AI Competition}

The  first Text-Based Adventure AI Competition\footnote{\url{http://atkrye.github.io/IEEE-CIG-Text-Adventurer-Competition/}.}, organized by the group from the University of York, has been announced in May 2016 and took place at the 2016 IEEE CIG conference in September. The second, will be held this year, also co-located with CIG.

The purpose of the competition is to stimulate research towards the transcendent goal of creating General Problem Solver, the task stated nearly six decades ago \cite{Newell1959Report}. The organizers plan to gradually increase the level of given challenges, with time providing more complex games that require more sophisticated approaches from the competitors. Thus, finally force them to develop agents that can autonomously acquire knowledge required to solve given tasks from the restricted domain of text-based games.

The domain of the competition is specified as any game that can be described by the Z-machine, the classic text adventuring engine. Interactive Fiction games are distributed as compiled byte code, requiring the special interpreter to run them. The first Z-machine has been developed in 1979 by Infocom, and supports games developed using a LISP-like programming language named Infocom's ZIL (Zork Implementation Language). The Text-based AI Competition uses Frotz\footnote{\url{http://frotz.sourceforge.net}.}, the modern version of the Z-machine, compatible with the original interpreter.

The competition organizers provide a Java package managing the communication between a game file and an agent process. Also, example random agents in Java and Python 3 are available. The interpreter is extended by the three additional commends, allowing players to quit the game, restart it  with a new instance of the agent, and restart without modifying the agent. Given that, the text-based AI framework supports learning and simulation-based approaches.

Little details about the competition insides are available. In particular, the number of participants is unknown, and the test environment game used to evaluate agents remained hidden, as it is likely to be used again this year. The game has been developed especially for the purpose of the competition and supports graduated scale of scoring points, depending on the quality of agent's solution.

The winner of the first edition was the BYU-Agent\footnote{The agent is open source and available at \url{https://github.com/danielricks/BYU-Agent-2016}.} from the Perception Control and Cognition lab at Brigham Young University, which achieved a score 18 out of 100. The idea behind the agent has been described in \cite{Fulda2017WhatCan}. It uses Q-learning \cite{Watkins1992Q} to estimate the utility of an action in a given game state, identified as the hash of its textual description.

The main contribution concerns affordance detection, used to generating reasonable set of actions. Based on the word2vec \cite{Mikolov2013Efficient}, an algorithm mapping words into a vector representations based on their contextual similarities, and the Wikipedia as the word corpus, the verb-noun affordances are generated. Thus, the algorithm is able to detect, for an in-game object, words with a similar meaning, and provide a set of actions that are possible to undertake with that object.

Provided results, based on the IF games compatible with Z-machine interpreter, shows the overall ability of the algorithm to successfully play text-based games. Usually, the learning process results in increasing score, and requires playing a game at least 200 times to reach the peek. However, there are some games that achieve that point much slower, or even the score drops as the learning continues.

\section{The Game Playing Agent}

\subsection{Overview}

Our agent is characterized by the following features.:

\begin{itemize}
\item it uses a huge set of predefined command patterns, obtained by analyzing various domain-related sources; the actual commands are obtained by suitable replacements;
\item it uses language models based on selection of fantasy books;
\item it takes advantage of the game-specific behaviors, natural for adventure games, like fight mode, equipment management, movement strategy;
\item it memorizes and uses some aspects of the current play history;
\item it tries to imitate human behavior: after playing several games and exploring the game universe it repeats the most promising sequence of commands. We treat the result reached in this final trial as the agent's result in this game. 
\end{itemize}

The agent was named ``Golovin'', as
one of the first answers it gives after asking \emph{Hey bot, what is your name?}, was \emph{your name is Golovin}, a phrase from the game \texttt{Doomlords}. 

\subsection{Preprocessing}

\subsubsection{Language Models}

 We used language models for two purposes. First, they allow us to define words similarity (which in turns gives us opportunity to replace some words in commands with their synonyms). For this task we use word2vec \cite{Mikolov2013Efficient} (and its implementation in TensorFlow \cite{tensorflow2015-whitepaper}). Secondly, we use neural network language models to determine which words in the scene description plays more important role than other (and so are better candidates to be a part of the next command). We use the LSTM neural networks operating on words \cite{Bengio03aneural}, augmented by the attention mechanism (\cite{DBLP:journals/corr/RaffelE15} and \cite{DBLP:journals/corr/ChengDL16}). This combination was previously tested in \cite{praca-szymona}. 

Since the action of many games is situated in fantasy universe, we decided to train our models on the collection of 3000 fantasy books from bookrix.com (instead of using Wikipedia, as in \cite{Fulda2017WhatCan}). 
 
\subsubsection{Commands}

In order to secure out agent against overfitting, we fix the set of games used in tests (the same 50 games as in \cite{Fulda2017WhatCan}). No data somehow related to this games were used in any stage of preprocessing.

We considered three methods to gather commands:
\begin{itemize}
\item \textbf{walkthroughs} -- for several games, sequence of commands from winning path can be found in the Internet. This source provides raw valid commands, that are useful in some games.
\item \textbf{tutorials} -- on the other hand, some games have tutorials written in natural language. Analyzing such tutorials\footnote{This tutorials were downloaded from the following sites: \url{http://www.ifarchive.org/}, \url{https://solutionarchive.com/}, \url{http://www.gameboomers.com/}, \url{http://www.plover.net/~davidw/sol/}} seemed to be a promising way of getting valid command.  Concept of reading manuals has been successfully used to learn how to play Civilization II game \cite{branavan2012learning}.
\item \textbf{games} -- at the end, there are many games that don't have tutorials nor walkthroughs. We downloaded a big collection of games, decompiled their codes, and extracted all texts from them.
\end{itemize}

The last two sources required slightly more complicated preprocessing. After splitting texts into sentences, we parsed them using PCFG parser from NLTK package \cite{Bird:2009:NLP:1717171}. Since commands are (in general) verb phrases, we found all minimal VP phrases from parse trees. After reviewing some of them, we decided not to take every found phrase, but manually create the list of conditions which characterizes 'verb phrases useful in games'. In this way we obtained the collections of approximately 250,000 commands (or, to be more precisely, command patterns). We also remember the count of every command (i.e. the number of parse tree it occurs in).

Some of the commands have special tag: ``useful in the battle''. We have manually chosen five verbs, as the most commonly related to fighting: \emph{attack}, \emph{kill}, \emph{fight}, \emph{shoot}, and \emph{punch}. Approximately 70 most frequent commands containing one of these verbs received this tag. 

The commands used by our agent were created from these patterns by replacing (some) nouns by nouns taken from the game texts. 


\subsection{Playing Algorithm}
The algorithm uses 5 types of command generators: battle mode, gathering items, inventory commands, general actions (interacting with environment), and movement. The generators are fired in the given order, until a non-empty set of commands is proposed.

There are multiple reasons why some generator may not produce any results, e.g.\
the agent reaches the predefined limit of making actions of that type, all the candidates are blacklisted, or simply we cannot find any appropriate command in the database. We will describe other special cases later.

When the description of the area changes, all the command lists are regenerated.

\subsubsection{Generating Commands}
Our general method to compute available commands and choose the one which is carried out, looks as follows:

\begin{enumerate}
\item Find all nouns in the state description and agent's equipment. (We accept all type of nouns classified by the \texttt{nltk.pos\_tag} function.)
\item Determine their synonyms, based on the cosine similarity between word vectors. (We use $n$-best approach with $n$ being subject to Spearmint optimization; see~\ref{sec:best_agent}.)
\item Find the commands containing nouns from the above described set. If a command contains a synonym, it is replaced by the word originally found in the description.
\item Score each command taking into account:
\begin{itemize}
\item cosine similarity between used synonyms and the original words
\item uniqueness of the words in command, computed as the inverse of number of occurrences in the corpora.
\item the weight given by the neural network model
\item the number of words occurring both in the description and in the command
\end{itemize}
The score is computed as the popularity of the command (number of occurrences in the command database) multiplied by the product of the above. The formula uses some additional constants influencing the weights of the components.
\item Then, using the score as the command's weight, randomly pick one command using the roulette wheel selection.
\end{enumerate}

\subsubsection{Battle Mode}

The battle mode has been introduced to improve the agent's ability to survive. It prevents from what has been the main cause of agent's death before -- careless walking into an enemy or spending too much time searching for the proper, battle-oriented and life-saving, action.

The agent starts working in battle mode after choosing one of the ``fight commands''. Being in this mode, the agent strongly prefers using battle command, moreover it repeats the same command several times (even if it fails), because in many games the opponent has to be attacked multiple times to be defeated. Therefore, between the consecutive fighting actions we prevent using standard commands (like \emph{look}, \emph{examine}), as wasting precious turns usually gives the opponent an advantage.

\subsubsection{Inventory Management (gathering and using items)}


In every new area, the algorithm searches its description for interesting objects. It creates a list of nouns ordered by the weight given by the neural network model and their rarity. Then, the agent tries \emph{take} them.

If it succeeds (the content of the inventory has changed), a new list of commands using only the newly acquired item is generated (using the general method). The constant number of highest scored commands is immediately executed.

\subsubsection{Exploration}

The task of building an IF game map is difficult for two reasons. One, because a single area can be presented using multiple descriptions and they may change as the game proceeds. Two, because there may be different areas described by the same text. Our map building algorithm tries to handle these problems.

We have found that usually the first sentence of the area description remains unchanged, so we use it to label the nodes of the graph (we have tried other heuristics as well but they performed worse). This heuristic divides all visited nodes into the classes suggesting that corresponding areas may be equivalent. The edges of the graph are labeled by the move commands used to translocate between the nodes (we assume that movement is deterministic). 

We initialize the map graph using the paths corresponding to the past movements of the agent. Then, the algorithm takes all pairs of nodes with the same label and considers them in a specific, heuristic-based, order. For every pair, the MergeNodes procedure (Listing~\ref{alg:mapping_merge}) is fired. The procedure merges two states joining their outcoming edges and recursively proceeds to the pairs of states that are reachable using the same move command. If the procedure succeeds, we replaces current map graph with the minimized one, otherwise the changes are withdrawn.

\begin{algorithm}\caption{MergeNodes($A$, $B$)}\label{alg:mapping_merge}
\begin{algorithmic}[1]
\AState {}\IIf{$A=B$} \textbf{return} \emph{True} \EndIIf \\
\IIf{label($A$)$\neq$label($B$)} \textbf{return} \emph{False} \EndIIf 
\AState \emph{mergelist}  $\gets$ $\{\}$
\ForAll{$m \in MoveCommands$}
  \If{$A$.moveby($m$)$\neq$\emph{None} $\wedge$ $B$.moveby($m$)$\neq$\emph{None}} 
    \AState{$mergelist$.append($(A$.moveby($m$)$, B$.moveby($m$)))}
  \EndIf
\EndFor
\AState JoinIncomingAndOutgoingEdges($A$, $B$)
\ForAll{$(A', B') \in mergelist$}\\
  \IIfa{$\neg$ MergeNodes($A'$, $B'$)} \textbf{return} \emph{False} \EndIIf 
\EndFor
\AState \textbf{return} \emph{True}
\end{algorithmic}
\end{algorithm}


We use a small fixed set of movement commands (\emph{south}, \emph{northwest}, \emph{up}, \emph{left}, etc.) to reveal new areas and improve the knowledge about the game layout. When the agent decides to leave the area, it tries a random direction, unless it already
discovered all outgoing edges -- then it uses map to find a promising destination. We evaluate destination nodes minimizing the distance to that node plus the number of tested commands divided by the node's curiosity (depending on scores of available commands and untested movement commands). Then, the agent follows the shortest path to the best scored destination.

\subsubsection{Failing Commands} 
When, after executing a command, the game state (description) remains unchanged, we assume the command failed. Thus, the algorithm puts it on a blacklist assigned to the current location. The command on the location's black list is skipped by the command generators.

After any change in the agent's inventory, all blacklists are cleared.

\subsubsection{Restarts}
The Frotz environment used for the contest allows to restart the game, i.e.\ abandon current play and start again from the beginning. 

Me make use of this possibility in a commonsense imitating of the human behavior. When the agent dies, it restarts the game and, to minimize the chance of the further deaths, it avoids repeating the last commands of his previous lives. The agent also remembers the sequence of moves that lead to the best score and eventually repeats it. The final trial's result is used as the agent's result in the game.
 


\section{Experiments}

Our experiments had two main objectives: creating the most effective agent, and analyze how some parameters influence the agents performance. 

The most natural way to measure the agent performance is to use the score given by the game (divided by the maximum score, when we want to compare different games). However, there are many games in which our agent (as well as BYU-Agent) has problems with receiving non zero points. So, we have decided to reward any positive score and add to the positive agent result arbitrarily chosen constant $0.2$. Therefore, optimal (hypothetical) agent would get $1.2$ in every game.

We selected 20 games for the training purposes, for all of them the maximum score is known. The performance of the agent is an averaged (modified) score computed on these games.

\subsection{Creating The Best Agent}
\label{sec:best_agent}
The agent's play is determined by some parameters, for instance:

\begin{itemize}
  \item the set of command patterns,
  \item the number of synonyms taken from word2vec,
  \item the number of items, we are trying to gather, after visiting new place, 
  \item the number of standard command, tried after gathering phase,
  \item how to reward the commands containing many words from description (the actual reward is $b^k$, where $k$ is the number of common words, and $b$ is a positive constant),
  \item how to punish the commands containing words with no good reason to be used (neither in state description nor in generated synonyms), 
      the score is divided by $p^k$, where $k$ is the number of such words, and $p$ is a constant,
  \item how many command should be done before trying movement command.
\end{itemize}

Furthermore we wanted to check, whether using battle mode or a map has an observable effect on agent performance. The number of parameter combinations was too large for grid search, so we decided to use Spearmint\footnote{Spearmint is a package which performs Bayesian optimization of hyperparameters. It allows to treat the optimized function as a black-box, and tries to choose the parameters for the next run considering the knowledge gathered during previous runs. See \cite{NIPS2012_4522}.}.

We started this optimization process with (total) score equal to 0.02, and after some hours of computation  we end with 0.08 (which means that the score has been multiplied 4 times). From now all parameters (if not stated otherwise) will be taken from the final Spearmint result.

\subsection{Evaluation of Domain-based Mechanisms}

\begin{figure}[!ht]
\centering
\includegraphics[height=50mm]{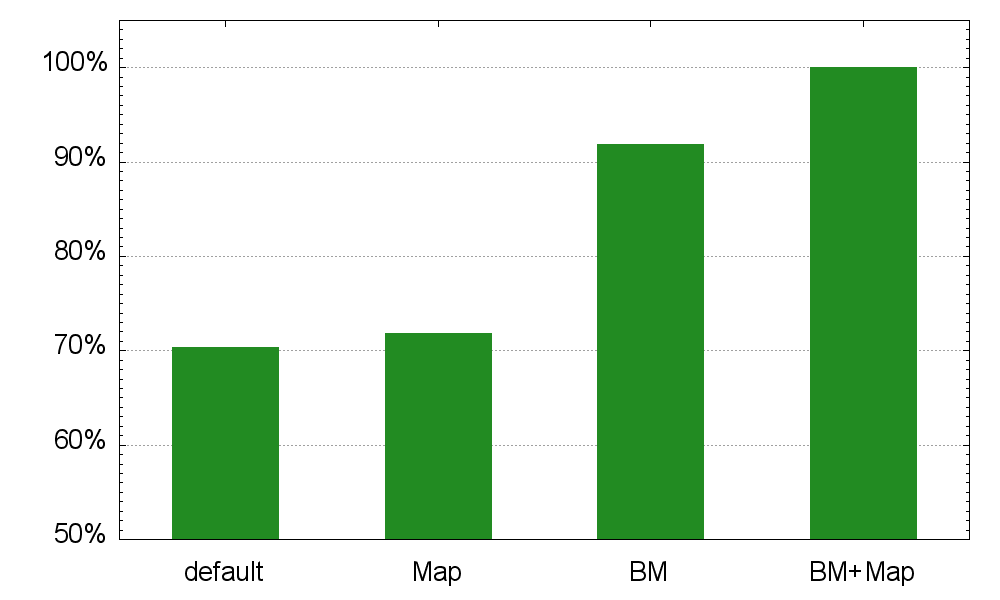}
\caption{Comparison of agent version with and without map and battle mode. Best variant scaled to 100\%.}\label{fig:domainbased}
\end{figure}

We wanted to check whether some more advanced features of our agent give observable influence on agent performance. We checked the following 4 configurations with battle-mode and map turned on or off. The results are presented in Figure \ref{fig:domainbased}. One can see that map is useful (but only to some extent), and battle mode is undoubtedly useful.

\subsection{Evaluation of Language Model Sources}

\begin{figure}[!ht]
\centering
\includegraphics[height=50mm]{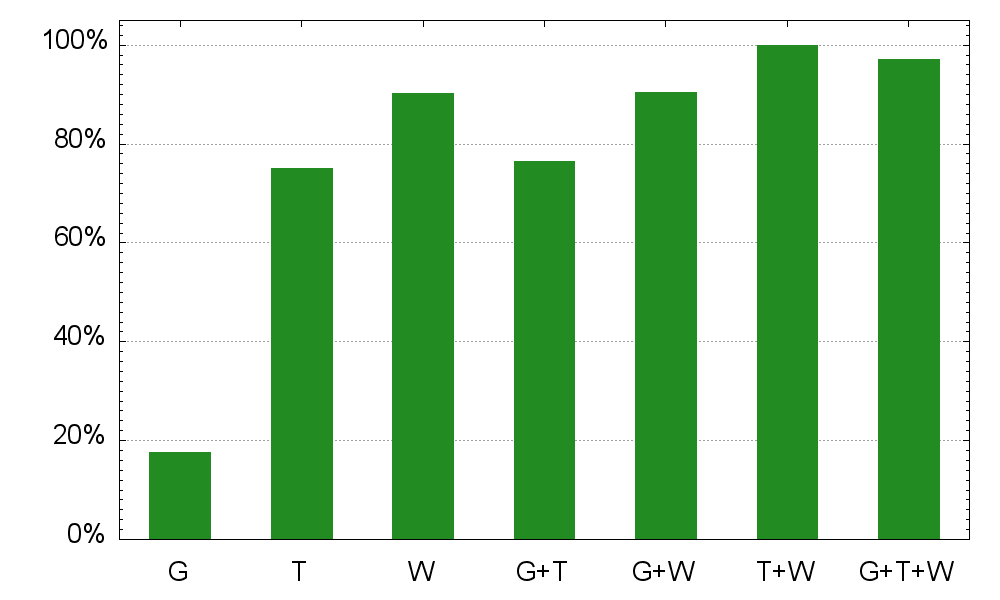}
\caption{Comparison of agent using different sources of commands. Best variant scaled to 100\%.}\label{fig:commsources}
\end{figure}

Commands were taken from 3 sources: tutorials (T), walkthroughs (W), and state description from games (G). We compared the agents used command from all combination of these sources. The results are presented in Figure \ref{fig:commsources}. The optimal configuration uses only two sources: T and W\footnote{The difference between T+W and G+T+W is not very big. In the previous version of this experiment the winner was G+T+W.}. We, however, still believe that decompiled games can be a useful source for game commands. But they cannot be found in descriptions, but in command interpreter -- which requires more advanced automated code analysis. We left it as a future work.

%


\subsection{Gameplay Examples}
\label{sec:example_detective}
While playing \texttt{detective}, our agent finds himself in a closet. We get the following state description:
\begin{quote}
\em \textbf{Game:} You are in a closet. There is a gun on the floor. Better get it. To exit, go east.
\end{quote}
Our agent determines items: \emph{closet}, \emph{gun}, \emph{floor}, \emph{exit}. Our agent is choosing from the commands listed in Table~\ref{tab:example_detective}. We see that what is the most important for the agent is gun and how to take it, which is reasonable and even suggested by the game. Moreover, the agent also tries to search using synonyms of the word \emph{gun} in order to find proper commands (for instance: we can see that he recognizes \emph{gun} as some kind of weapon, and some weapons, as for example knife, can be \emph{sharpened}). Fortunately, after using a command \emph{get gun}, Golovin obtains a black little pistol.

\begin{table}[htpb]
\caption{Best 10 (out of 25) commands proposed by our agent for the situation described in the \texttt{detective} example (Section~\ref{sec:example_detective})}
\label{tab:example_detective}
\centering
\begin{tabular}{c|r}
action & score \\ \hline
\emph{get gun}      & 0.1736 \\  
\emph{drop gun}        & 0.1129 \\ 
\emph{take gun}     & 0.0887 \\ 
\emph{open closet}     &  0.0557 \\ 
\emph{examine gun}   & 0.0309 \\
\emph{fire gun}       & 0.0252 \\ 
\emph{load gun}         & 0.0237 \\ 
\emph{examine closet}   & 0.0128 \\
\emph{buy gun}    &  0.0042 \\
\emph{sharp gun}  & 0.0006 \\
\end{tabular}
\end{table}

Another example comes from \texttt{zork3}. We start with the following location description:

\begin{quote}
\em \textbf{Game:} You are in a shadowy land of low, rolling hills stretching out to the west and south. The land is bordered to the north by a massive stone wall. Ancient and weathered, the wall has crumbled enough at one point to permit passage. A cloaked and hooded person, carrying a sword not unlike your own, is standing blocking the way to the northeast. The hooded figure has a great deal of strength, perhaps matching your own.\\
\textbf{Golovin:} examine figure\\
\textbf{Game:} There is nothing special to be seen. Your quick reflexes cannot stop the hooded figure's stroke! You are hit! 
\end{quote}


In this very moment, in the agent's inventory are: \emph{a sword}, \emph{a lamp}.  
In the command list we can find for instance \textit{kill man with sword}. As the word \emph{man} is close enough to \emph{figure}, and Golovin also have \emph{sword}, he chooses to use command \textit{kill figure with sword}. Because the fighting command was chosen, the battle mode is turned on, and the game continues as follows:
\begin{quote}
\em 
\textbf{Golovin:} kill figure with sword\\
\textbf{Game:} The hooded figure is hit with a quick slash!
The figure has a light wound which hasn't affected its seemingly great strength.
\end{quote}
the agent's score improves and the fight continues\ldots 
\begin{quote}
\em 
\textbf{Golovin:} kill figure with sword\\
\textbf{Game:} You charge, but the hooded figure jumps nimbly aside. The hooded figure stabs nonchalantly with its sword and misses. 
\end{quote}

\subsection{The Comparison with BYU-Agent}

Finally, we validate our approach by comparing it with the BYU-Agent. We were using the same set of 50 Z-machine games\footnote{The game set is available at \url{https://github.com/danielricks/textplayer/tree/master/games}.} as in \cite{Fulda2017WhatCan}.

The results of the comparison are presented in Table~\ref{tab:byu_comparison}. The BYU agent was trained for 1000 epochs (each epoch containing 1000 game steps), and its score was noted after each epoch. Because the learning curves vary depending on the game, including degeneration of the results  (see \cite[Figure~5]{Fulda2017WhatCan}), as the main measure we took the maximum score achieved over all epochs. 

As for the Golovin, we restricted his playing time to 1000 steps (i.e.\ an equivalent of one epoch) and use our commonsense restarting mechanism.

The BYU-Agent results are obtained using the verb and action space reduction algorithm, except the games marked with an asterisk, where the verb space reduction experienced errors, so we present scores obtained by the action space reduction variant instead.

Eventually, there are 24 games, out of 50, where some of the agents received any positive reward. Golovin scored better in 12 games, including 7 games where BYU-Agent received no reward. BYU-Agent scored better in 11 games, including 6 games where Golovin scored no points. One game is a non-zero tie.

Thus, despite significantly shorter learning time (i.e.\ available number of steps), our agent is able to outperform BYU-Agent on a larger number of games than he is outperformed on. On the other hand, BYU-Agent gains in the games where the Q-learning is effective and gradually increases score through the epochs, e.g.\ \texttt{curses}, \texttt{detective} or \texttt{Parc}.

Last observation concerns the number of games where only one of the agents scored 0, which is surprisingly large. This may suggest that the two compared approaches are effective on a different types of games, and may, in some sense, complement each other.

\begin{table}[htpb]
\caption{Average scores for 10 runs of each game. For BYU-Agent we took the maximum achieved score during the 1000 epochs training. Golovin plays for one epoch. In the games that are not listed both agents gain no reward. The asterisk marks games that uses other version of BYU-Agent}
\label{tab:byu_comparison}
\centering
\begin{tabular}{c|rr|r}
game               & \quad Golovin  & BYU-Agent     & max score\\ \hline
\texttt{balances}  & \textbf{9.0}    & 0              & 51 \\
\texttt{break-in}  & 0             & \textbf{0.3}   & 150 \\
\texttt{bunny}     & \textbf{2.7}             & 2.0   & 60 \\
\texttt{candy}     & 10.0             & 10.0  & 41 \\
\texttt{cavetrip}  & \textbf{15.0}             & 10.5  & 500 \\
\texttt{curses}    & 0.4             & \textbf{1.9}   & 550 \\
\texttt{deephome}  & \textbf{1.0}    & 0              & 300 \\
\texttt{detective} & 71.0             & \textbf{213.0} & 360 \\
\texttt{gold} & \textbf{0.3} & 0 & 100 \\
\texttt{library}   & \textbf{5.0}    & 0              & 30 \\
\texttt{mansion}   & 0.1             & \textbf{2.2}   & 68 \\
\texttt{Murdac}    & \textbf{10.0}    & 0              & 250 \\
\texttt{night} & \textbf{0.8} & 0 & 10 \\
\texttt{omniquest} & \textbf{7,5}             & 5.0   & 50 \\
\texttt{parallel}  & 0             & \textbf{5.0}   & 150 \\
\texttt{Parc}      & 1.6             & \textbf{5.0}   & 50 \\
\texttt{reverb}    & 0             & \textbf{1.8}   & 50 \\
\texttt{spirit}    & \textbf{3.2}          & 2.0  & 250 \\
\texttt{tryst205}  & 0.2             & \textbf{2.0}   & 350 \\
\texttt{zenon}     & 0               & \textbf{2.8}   & 20 \\
\texttt{zork1}     & \textbf{13.5}   & *8.8  & 350 \\
\texttt{zork2}     & -0.1            & *\textbf{3.3}  & 400 \\
\texttt{zork3}     & \textbf{0.7}    & *0  & 7 \\
\texttt{ztuu}      & 0               & \textbf{0.5}   & 100 \\
\hline
better in: & 12 & 11 & games
\end{tabular}
\end{table}

\section{Conclusions and Future Work}

We have presented an agent able to play any interactive fiction game created for human players, on the level comparable to the winner of the last year Text-Based Adventure AI Competition. Due to the number of domain-based mechanisms, our agent can successfully handle the game in a limited number of available steps. The results of the presented experiments show that the mechanisms we embed (battle mode, mapping) and a choice of learning sources, indeed improves the agent's performance.

Although the results are promising, we are still at the beginning of the path towards creating the agent that can really understand the natural language descriptions in order to efficiently play the text-based adventure games.

There are multiple future work directions we would like to point out. First, and one of the most important, is to embed a learning mechanisms: the in-game learning, that uses restart functionality to improve player efficiency in one particular game; and preliminary learning, that is able to gain useful knowledge from playing entire set of games. Also, we plan to take a closer look at the decompiled game codes, as we believe that analyzing them may provide very useful knowledge.

We would like to improve the battle mode behavior, especially mitigate the agent and make it more sensitive to the particular situation. We hope that the mapping mechanism can be further extended to allow the casual approach, where the agent travels to distant locations for some specific reason (e.g.\ item usage), instead of a simple reactive system that we have now. 

Lastly, we would like to continue the domain-based approach, and so focus our efforts on discovering the subgames (like we did with fighting and exploring) that we are able to properly detect, and handle significantly better than the general case.

\section*{Acknowledgments}

The authors would like to thank Szymon Malik for his valuable contribution in the early stage of developing Golovin.

We would also like to thank Nancy Fulda for helpful answers to our questions and providing up-to date results of the BYU-Agent.

\bibliographystyle{IEEEtran}
\bibliography{textbasedai}

\end{document}